\documentclass[letterpaper, 10 pt, journal, twoside]{IEEEtran}
\usepackage{dblfloatfix} 
\usepackage{hyperref}
\usepackage[all]{hypcap}
\usepackage[utf8]{inputenc}
 
\usepackage{multicol}
\usepackage{tikz}
\usetikzlibrary{arrows}
\usepackage{makecell}
\usepackage{booktabs}
\usetikzlibrary{positioning,fit}
\usetikzlibrary{arrows.meta}
\usepackage{adjustbox}
\usepackage[cmex10]{amsmath}
\usepackage{amssymb}
\usepackage{colonequals}
\usepackage{pgfplots}
\pgfplotsset{compat=1.14}
\usepackage{etoolbox}
\usepackage[ruled, vlined]{algorithm2e}
\usepackage{cleveref}

\SetKwInput{KwPersistent}{Persistent variables}
\SetKwInput{KwLocal}{Local variable initializations}
\SetArgSty{textnormal}
\makeatletter
\newcommand{\nosemic}{\renewcommand{\@endalgocfline}{\relax}}
\newcommand{\dosemic}{\renewcommand{\@endalgocfline}{\algocf@endline}}
\newcommand{\pushline}{\Indp}
\newcommand{\popline}{\Indm\dosemic}
\makeatother

\makeatletter
\newcommand{\removelatexerror}{\let\@latex@error\@gobble}
\makeatother

\makeatletter
\newcommand*{\hyperlinkcite}[1]{\hyper@link{cite}{cite.#1}}
\makeatother

\newcommand*{\vcenteredhbox}[1]{\begingroup
\setbox0=\hbox{#1}\parbox{\wd0}{\box0}\endgroup}
\newcommand*{\mat}[1]{\mathbf{#1}}

\usepackage{todonotes}
\usepackage[normalem]{ulem}
\usepackage{color, soulutf8}
\begin{document}

\setcounter{page}{0}

\onecolumn
\pagenumbering{gobble}

\noindent © 2019 IEEE. Personal use of this material is permitted. Permission from IEEE must be obtained for all
other uses, in any current or future media, including reprinting/republishing this material for advertising
or promotional purposes, creating new collective works, for resale or redistribution to servers or lists, or
reuse of any copyrighted component of this work in other works. \\

\noindent Journal: IEEE Robotics and Automation Letters \\
DOI: 10.1109/LRA.2019.2928203 \\
URL: \url{https://ieeexplore.ieee.org/document/8760392}

\newpage
\twocolumn
\pagenumbering{arabic}

%
\title{Efficient Dense Frontier Detection for 2D Graph SLAM Based on Occupancy Grid Submaps}

\markboth{IEEE Robotics and Automation Letters. Preprint Version. Accepted June, 2019}
{Oršulić \MakeLowercase{\textit{et al.}}: Efficient Dense Frontier Detection for 2D Graph SLAM Based on Occupancy Grid Submaps}  

\author{Juraj Oršulić$^{1}$, Damjan Miklić$^{2}$ and Zdenko Kovačić$^{1}$
\thanks{Manuscript received: February 24, 2019; revised May 23, 2019; accepted June 25, 2019. This paper was recommended for publication by Editor Eric Marchand upon evaluation of the Associate Editor and Reviewers' comments.} 
\thanks{$^{1}$The authors are with Faculty of Electrical Engineering and Computing, Laboratory for Robotics and Intelligent Control Systems, University of Zagreb, Croatia.  Corresponding author: {\tt\footnotesize juraj.orsulic@fer.hr}}
\thanks{$^{2}$The author is with RoMb Technologies d.o.o., Zagreb, Croatia.}
\thanks{Digital Object Identifier (DOI): see top of this page.}
}

\maketitle

\begin{abstract}
In autonomous robot exploration, the frontier is the border in the world map dividing the explored and unexplored space. The frontier plays an important role when deciding where in the environment the robots should go explore next. We consider a modular control system pipeline for autonomous exploration where a 2D graph SLAM algorithm based on occupancy grid submaps performs map building and localization, and frontier detection is one of key system components. We provide an overview of the state of the art in frontier detection and the relevant SLAM concepts and propose a fast specialized frontier detection method which is efficiently constrained to active submaps, yet robust to graph SLAM loop closures.
\end{abstract}

\begin{IEEEkeywords}
Visual-Based Navigation; Computer Vision for Other Robotic Applications; Path Planning for Multiple Mobile Robots or Agents
\end{IEEEkeywords}


\tikzstyle{block} = [draw, fill=blue!20, rectangle, text centered, text width=12em,
    minimum height=3em, minimum width=6em, node distance = 1cm]
\tikzstyle{vblock} = [block, node distance = 0.5cm]
\def\edgedist{2.5}

\pgfdeclarelayer{background}
\pgfdeclarelayer{foreground}
\pgfsetlayers{background,main,foreground}

\begin{figure}[b]\setlength{\hfuzz}{1.1\columnwidth}
\begin{minipage}{\textwidth}
\centering
\begin{tikzpicture}[auto, node distance=1cm,>={Latex[length=2mm]}]
    \node [block, name=acquisition] {Sensor data acquisition \\ (lidar, IMU, odometry) };
    \node [block, name=slam, right=of acquisition, yshift=0.5cm] [block] {SLAM};
    \node [vblock, name=detection, fill=red!20, below=0 cm
    of slam] {Frontier detection};
    \node [block, name=exploration, right=2 cm of slam, yshift=0.25cm]{Exploration task \\ generation and scheduling};
    \node [vblock, name=planning, below=of exploration]{Path planning \\ and following};
     \draw[->] (exploration) -- (planning);
     \path (planning.south)+(0,-0.7) node[coordinate, name=feed1, inner sep=25pt] {};
     \path (planning.south -| acquisition.south)+(+0,-0.7) node[coordinate, name=feed2, inner sep=0pt] {};
     \draw[->, dashed]  (planning.south) -- (feed1) -- node {Environment} (feed2) -- (acquisition);

    \begin{pgfonlayer}{background}
        \path (detection.south -| detection.west)+(-0.2,-0.2) node (a) {};
        \path (slam.north -| slam.east)+(0.2,0.2) node (b) {};
        \node (f1)  [draw=black!50, fill=yellow!20, inner sep=0pt, 
             rounded corners, dashed,
             fit=(a) (b),
             name=rectangle] {};
    \end{pgfonlayer}
    
    \path (acquisition.east -| rectangle.west) node[coordinate, name=rectleft] {};
    \path (rectangle.east -| exploration.west)+(-0.8,0) node[coordinate, name=rectright] {};
    \draw[->] (acquisition.east) -- (rectleft);
    \draw (rectangle.east) -- (rectright);
    \draw[->] (rectright) |- (exploration.west);
    \draw[->] (rectright) |- (planning.west);
\end{tikzpicture}
\caption{Block diagram providing a high-level overview of the considered control system for performing autonomous exploration. The robot may be equipped with a laser rangefinder, an inertial measurement unit, and wheel encoders for odometry. A Simultaneous Localization and Mapping module uses the sensed environment data to build an environment map and to estimate the pose therein, while a frontier detection module keeps track of the exploration frontier. An exploration task generation and scheduling module assigns exploration tasks to be performed according to an exploration strategy and forwards it to a path planning and following module, which uses the map and localization from SLAM to steer the robot towards the goal defined by the assigned task. Notably, the frontier detector is tightly coupled with the SLAM module, enabling efficient implementation of frontier detection.}
\label{fig:high_level_block_diagram}
\end{minipage}
\end{figure}
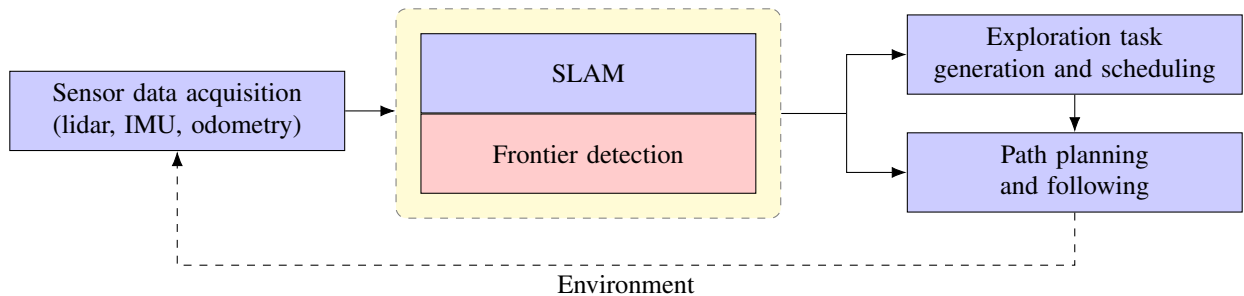

%
\IEEEpeerreviewmaketitle

\section{Introduction}

\IEEEPARstart{A}{} fully autonomous mobile robot, able to explore, navigate and perform actions in an unknown environment is one of the ultimate objectives of today's mobile robotics research. To this end, we consider a single autonomous robot or an autonomous team of robots tasked with exploring the unknown environment. The autonomous exploration problem comprises collecting the data sensed from the environment, using the collected data to build a structured model of the environment, self-localizing in the environment model, high-level planning and scheduling of robot tasks (mission generation and assignment), and path planning and following. All of these need to be performed in real time.

In a common approach known as frontier exploration, the robot maintains information about the border which divides the explored and unexplored space in the environment -- the \emph{frontier}. Elements of the frontier represent places in the environment which the robot may approach and thereby increase the knowledge about the structure of the environment. With the information about the exploration frontier available,  mission planning can be described in its simplest version as \emph{(boldly) go where no one has gone before}.

Many components of the autonomous exploration problem are complex enough to be associated with their own field of robotics research, resulting in sophisticated methods and software modules being available for solving them. Namely, building the model of the environment (a map) and self-localization therein may be performed by a module implementing a Simultaneous Localization and Mapping (SLAM) algorithm.
For this reason, we consider a control system implementing a modular exploration pipeline as depicted in \autoref{fig:high_level_block_diagram}. Map building and localization are performed by a SLAM module. The map built by SLAM is processed in the frontier detection module, and finally the map, the localization pose and the detected frontier are used by the later stages of the pipeline (exploration task generation and scheduling, path planning and following) to guide the robot according to an exploration strategy.

\enlargethispage{-16.5\baselineskip}
The scope and the main contribution of this paper is the frontier detection part of the exploration pipeline -- a new, efficient frontier detection approach, specialized for use with a 2D graph SLAM algorithm based on occupancy grid submaps \cite{hess2016}. The proposed approach is robust to loop closures and exploits the submap structure of the SLAM algorithm in order to quickly perform frontier updates. By providing high frequency incremental frontier updates which enable more responsive planning of exploration objectives, a real-time use case is facilitated on large and complex maps, e.g. the Deutsches Museum dataset \cite{hess2016}. All the while the proposed frontier detection algorithm delivers a result at least as good as a naive frontier edge-detection algorithm, i.e. performing edge detection on a completely assembled global map each time after SLAM updates the map by inserting scans or optimizing the pose graph to perform loop closures.

\section{Related Work} \label{sec:related_work}

\subsection{Frontier Exploration as a Prevalent Exploration Method}

Frontier exploration in the context of autonomous robotics was first introduced by Yamauchi in 1997 \cite{yamauchi1997}, paving the way for many others (\cite{burgard2000, freda2005, wettach2010}).
 Commonly, elements of the detected frontier are used as navigation goals during planning of exploration tasks. Building on this, there are more complex exploration strategies which attempt to coordinate entire robot teams (\cite{burgard2005,faigl2013}),
or use frontiers as sinks in a potential field (\cite{shade2011}). Frontier detection is therefore a key elementary operation in frontier exploration, and it is important that it be performed as quickly as possible so that exploration can be more efficient \cite{quin2014}.

\subsection{State of the Frontier Detection Art}

A naive algorithm for frontier detection is to perform edge detection on the complete global map after each map update. However, this approach is not feasible for larger maps and real-time robot operation with such maps, as it presents a significant computational burden.

\subsubsection{Keidar and Kaminka's seminal work on efficient frontier detection}

Keidar and Kaminka \cite{keidar2014} proposed in 2014 several approaches which attempt to perform frontier detection in an efficient manner. The first, \emph{Wavefront Frontier Detector} (WFD), consists of running two consecutive breadth-first searches (BFS). The first BFS starts at the robot position and continues throughout the unoccupied space, until eventually a frontier point is found which belongs to a component of connected frontier points. From there, the rest of the connected component is found by a second BFS along the connected frontier points. While WFD avoids searching the unobserved space, it still searches all observed space in each iteration, which may degenerate into a full map search as exploration progresses.

The second approach to frontier detection proposed by Keidar and Kaminka, the \emph{Fast Frontier Detector} (FFD), does not use the map built by SLAM, but rather constructs the contour of each laser scan using Bresenham's line algorithm, and uses the constructed contour to detect the frontier and store it in a specialized data structure. Quin and Alempijević \cite{quin2014} note that FFD has to be executed after each scan, which results in many wasteful calculations if frontier updates are required only occasionally, and that Bresenham's line algorithm can cut across unobserved space and miss some frontier cells. Our proposed approach does not require execution after each processed scan, supporting a use case where frontier updates are required only occasionally.

FFD is also notable for introducing the concept of \emph{active area} -- a bounding box positioned in the map around the robot position, circumscribing the last scan the map was updated with. The frontier update step is sped up by restricting it to the active area. Keidar and Kaminka also applied this concept to the WFD detector, yielding the \emph{incremental WFD} (WFD-INC) algorithm, which requires non-trivial auxiliary data structures for frontier point maintenance. Our proposed algorithm has a similar concept of \emph{active submaps}.

\begin{figure*}
\smallskip
\adjustbox{max width=\textwidth}{
\vcenteredhbox{\includegraphics[width=.333\textwidth]{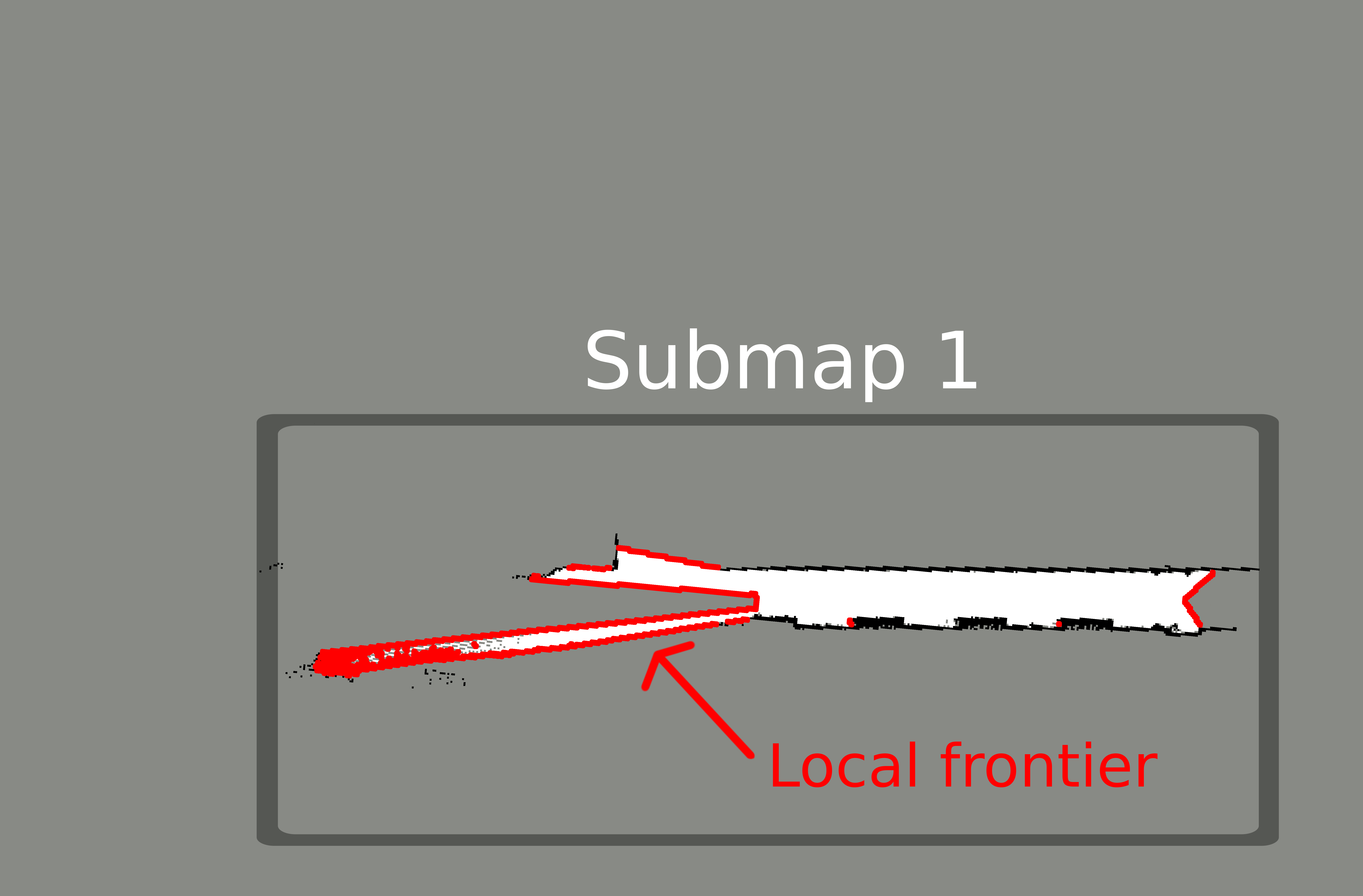}}
$\cup$
\vcenteredhbox{\includegraphics[width=.333\textwidth]{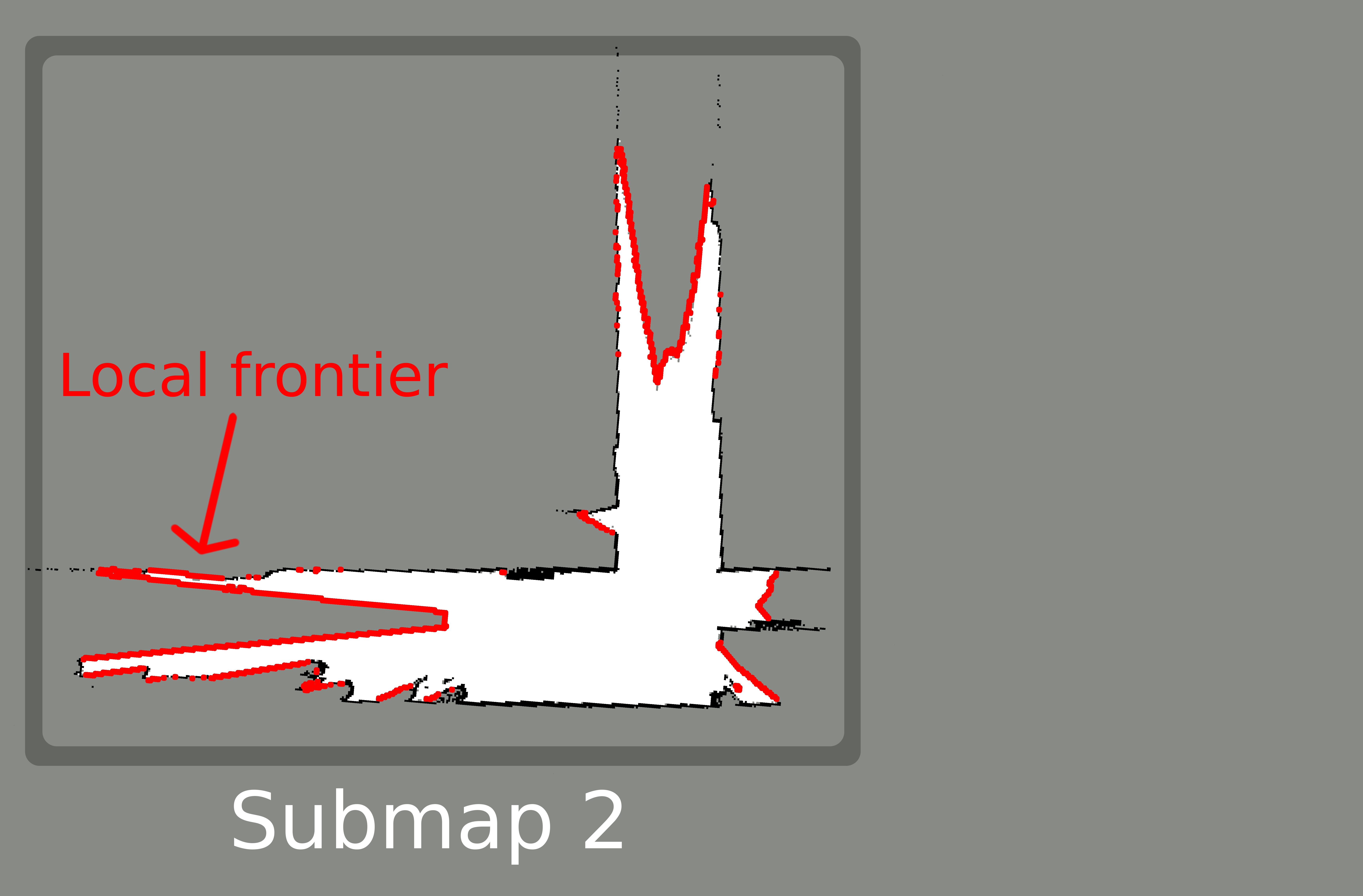}}
$=$
\vcenteredhbox{\includegraphics[width=.333\textwidth]{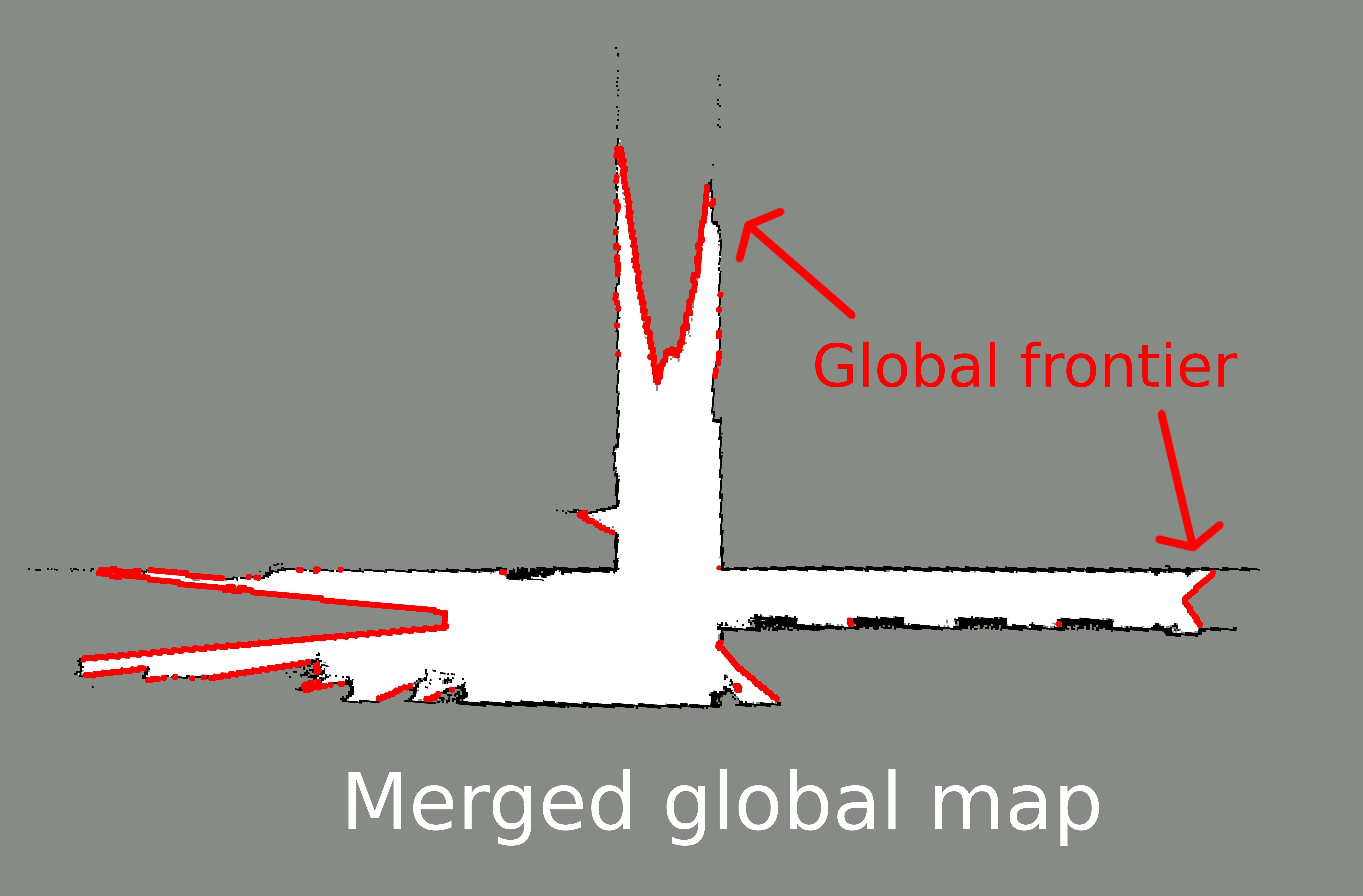}}
}
\caption{Depiction of merging two adjacent submaps and their local frontiers into the global composite map and the global frontier. Colour legend: unobserved grid cells are grey, unoccupied cells are white, occupied cells are black, while frontier points are red. Each submap has a \emph{local frontier} which is based solely on the occupancy grid of that submap. The points in the local frontier set are candidates for becoming part of the \emph{global frontier}. The submaps are then transformed into the global map coordinate system according to the current optimized solution of graph SLAM. Any local frontier point which after transforming ends up in an observed area of any other submap (i.e. fails the stabbing query test against that submap) is discarded from the global frontier set. An extended visualization of this figure is available in the video playlist linked in \autoref{sec:experimental_results}.}
\label{fig:submap_composition}
\end{figure*}

\subsubsection{Impact of loop closure in SLAM on frontier detection}

Loop closure is an event when the SLAM algorithm recognizes that the robot has revisited the same place, and then makes a correction using this information which reduces the error caused by drift in localization along the whole loop. In order to return a correct result, frontier detection has to be able to efficiently cope with the map changes induced by loop closure corrections. These map changes are not confined to the active area -- performing loop closure results in widespread changes all over the map. While an efficient frontier detection algorithm should avoid reassembling and iterating throughout the entire global map in every iteration, constraining the algorithm to only the active area makes it difficult to be robust to loop closures. WFD-INC addresses loop closure events by evicting the detected frontier and performing frontier detection from scratch using the new loop-corrected map.

To efficiently address loop closures, the frontier detection algorithm needs to get intimate 
to a certain degree with the implementation of the SLAM algorithm. For example, Keidar and Kaminka additionally proposed an implementation of WFD-INC for GMapping (a particle filter-based SLAM \cite{grisetti2007}) called \emph{incremental parallel} WFD (WFD-IP).
 WFD-IP performs in parallel separate WFD-INC frontier detection for each particle (each particle having its own map). 
 Like WFD-INC, our proposed method uses the internals of the SLAM algorithm in order to perform frontier detection faster while also being robust to loop closure.

Quin and Alempijević \cite{quin2014} introduce two frontier detection methods: \emph{naive active area} (NaiveAA), which is the naive approach confined to the active area, and a version of WFD called \emph{Expanding WFD} (EWFD) which steers the WFD breadth-first search into newly discovered unoccupied areas. EWFD assumes that the entropy for each cell can only decrease over time. This is not true in the general case for the complete global map when considering effects of loop closure -- observed areas can get moved around in the global map during loop closure and leave unexplored space in their wake. However, the entropy decrease assumption \emph{is} almost surely true for single submaps in submap-based SLAM, and we exploit this fact in our proposed approach.

\subsubsection{Other approaches}
Senarathne and Wang \cite{senarathne2013} use an oriented bounding-box based inexact approach.

Umari \cite{umari2017} uses rapidly-exploring random trees (RRT) to perform sparse frontier detection by building a tree inside the unoccupied space in the map built by SLAM. When the algorithm crosses the frontier while trying to expand the random tree, a single frontier point is detected. However, using the implementation of the algorithm provided in \cite{umari2017} does require reassembling the global map in each iteration. Also, this algorithm is not robust to loop closure, since the built RRT tree does not follow the results of pose graph optimization. In our experiments, we have observed various problems with performance of RRT frontier detection in narrow corridors and with large maps, prompting us to devise the method for dense frontier detection proposed herein.


\section{Prerequisites}

\subsection{Simultaneous Localization and Mapping}

The term SLAM was coined by Leonard and Durrant-Whyte in 1991 \cite{leonard1991}.
As shown in the block diagram of the exploration pipeline in \autoref{fig:high_level_block_diagram}, a SLAM algorithm uses sensor data to build a map and perform localization, which is further used in frontier detection, exploration task planning and execution. 
There is a wealthy trove of SLAM methods developed to this day, which can be roughly grouped into methods based on filtering and methods based on graph optimization. 
We will focus on graph SLAM, which represents poses and detected features as nodes in a graph, while the correspondences which impose constraints on the poses of the respective nodes are represented as graph edges. Various optimization methods may be used to minimize the residual error of all constraints, e.g. the Ceres solver \cite{ceres-solver}. 

Submaps are small local maps which are merged into a global map. One of earlier approaches to SLAM using submaps is \cite{williams2002}, with further examples being \cite{strom2011} and 
\cite{bosse2008}.

\subsection{Cartographer}
The proposed frontier detection method was designed for use with Cartographer, an open-source multi-robot multi-trajectory 2D and 3D graph SLAM based on occupancy grid submaps, developed by Google (Hess, Kohler, Rapp in 2016 \cite{hess2016}).
Cartographer's approach of optimizing the poses of all scans and submaps follows Sparse Pose Adjustment \cite{konolige2010} and uses the Ceres solver \cite{ceres-solver} for optimizing the pose graph using the Levenberg–Marquardt algorithm (LMA).

Submaps in Cartographer are spatially and temporally compact occupancy grid maps, typically of resolution 0.05 m, made from a short, continuous series of rangefinder sensor measurements (laser scans) taken during traversal of a short section of the robot trajectory. It is desired that the size of the submaps be small enough such that the localization drift is not perceptible within a single submap. 

Cartographer maintains a pair of \emph{active submaps} that the laser scans are inserted into, according to a \emph{local pose} obtained by performing scan matching against the older (more complete) submap from the active submap pair.

The occupancy probabilities of grid cells are initially unobserved i.e. unknown (exactly 0.5). When a predetermined fixed number of scans $n_{scans}$ is inserted into a submap, it is marked as finished, and a new submap is created to take its place in the active submap pair. Importantly, once a submap is finished, its occupancy grid is immutable from that point onward. 
Cell occupancy probabilities are clamped to the interval $[0.1, 0.9]$ and are stored linearly mapped onto the space of unsigned 16-bit integers. Scan insertion into submaps is performed as Bayesian updates of the cell occupancy probabilities (see (3) in \cite{hess2016}). The cells corresponding to laser hit points are updated with ``occupied'' observations, while the intermediate points (obtained by casting rays from the laser rangefinder origin to hit points) are updated with ``unoccupied'' observations.

It is important to note that on a level of a single submap, the cell entropy can be assumed to be monotonically decreasing, i.e. \emph{the cells of an active submap cannot become unobserved once they are observed}.

When Cartographer detects loop closures, 
pose constraints between the corresponding trajectory nodes and submaps are added as edges into the pose graph. Optimization is periodically invoked in order to find a new solution -- a set of global submap and trajectory node poses. 
As discussed, for frontier detection, this implies that when pose graph optimization is performed, the submaps can and do get displaced and rotated (i.e. undergo rigid transformations), although their occupancy grids are immutable after they are marked as finished. Our proposed frontier detection approach takes advantage of these properties of Cartographer.




\section{Frontier detection}

\subsection{Definitions}

\textbf{Rigid transformation} $\mat{T}^b_a \in \mat{SE}(3)$ is the pose of the coordinate system $b$ relative to the coordinate system $a$. Transforming a point with coordinates expressed in $b$, $\mat{p}^b \in \mathbb{R}^3$, into the corresponding point $\mat{p}^a$ in $a$ is denoted as $\mat{p}^a = \mat{T}^b_a \; \mat{p}^b$.

\textbf{The global map coordinate system} is denoted with $g$. The solution of pose graph optimization are poses of graph nodes expressed with respect to $g$. 



\textbf{Submap} is an occupancy grid with a \textbf{local submap coordinate system}, in which the submap occupancy grid cells are indexed with 2-integer tuples: $\mat{S}^{si}_{k, l}$ is the occupancy probability value of the cell $(k,l)$ in the submap $si$.  A submap is finished after being updated with $n_{scans}$ consecutive laser scans. \textbf{The set of active submaps} contains the submaps which are not yet finished.  

\textbf{Global submap pose} $\mat{T}_g^{si}$ is the global pose in $g$ of the origin of the local coordinate system of a submap $si$. Global submap poses are part of the optimized pose graph solution.

\textbf{Occupancy classification} -- the cell occupancy probability values are not used directly in frontier detection. The probability values are first classified according to the following thresholding rule:
\begin{equation}
     \text{class}(p = \mat{S}^{si}_{k, l}) \colonequals \begin{cases} 
      \text{unoccupied} & p < 0.5 \\
      \text{occupied} & p > 0.5 \\
      \text{unobserved} & p = 0.5
   \end{cases}
   \label{eq:thresholding}
\end{equation}

\textbf{Observed cells} are occupancy grid cells that are not unobserved. 

\textbf{Local frontier point} is the center of an  \textbf{unobserved} occupancy grid cell which is adjacent to a \textbf{unoccupied} cell in the same submap. \textbf{Local frontier} of a submap is the set of its local frontier points. The red points in the first two pictures in \autoref{fig:submap_composition} are examples of local frontiers.

\textbf{Stabbing query} refers to looking up the corresponding cell in a given submap for a given global point. More precisely, for a submap $si$ and a global point $\mat{p}^g$,
to find the occupancy grid cell $(k, l)$ in $si$ whose center coordinates in the local coordinate system of the submap $si$ are closest to $(\mat{T}_g^{si})^{-1} \mat{p}^g$. Further, performing a \textbf{stabbing query test} means checking if the corresponding cell $\mat{S}^{si}_{k, l}$ is an \textbf{unobserved} cell, in which case the test \textbf{passes}.

\textbf{Global frontier point} is the center of an \textbf{unobserved} global occupancy grid map cell adjacent to a \textbf{unoccupied} global map cell. A \textbf{valid} global frontier point must pass the stabbing query test against all submaps. \textbf{Global frontier} is the set of all valid global frontier points at a given time. The red points in the third picture in \autoref{fig:submap_composition} are an example of a global frontier.

\textbf{Perimeter} of the global or local frontier is the number of frontier points in the respective set.

It may be noted that the frontier points could alternatively have been defined as centers of \emph{unoccupied} cells adjacent to \emph{unobserved} cells. We have chosen to define frontier points as centers of \emph{unobserved} cells as above in order to simplify the stabbing query test. 

\capstartfalse
\begin{figure}[t!]
\removelatexerror
\begin{algorithm}[H]
 \small
 \capstart
 \KwIn{active\_submaps, global\_submap\_poses}
 \KwPersistent{\\ \quad local\_frontiers, global\_frontiers, \\ \quad global\_submap\_bounding\_boxes}
 \KwLocal{\\ \quad current\_bounding\_boxes $\Leftarrow$ empty
          \\ \quad submaps\_with\_frontier\_updates $\Leftarrow$ active\_submaps}
 \nl \ForEach{submap $si \in$ active\_submaps}{
 \nl local\_frontiers[$si$].Clear()\;
 \nl global\_frontiers[$si$].Clear()\;
  \nl \nosemic current\_bounding\_boxes[$si$] $\Leftarrow$\;
        \pushline CalculateGlobalBoundingBox($si$,\;
        \pushline\dosemic global\_submap\_poses[$si$])\;\popline\popline
  \lnl{algline:intersecting_submaps_query} \nosemic intersecting\_submaps $\Leftarrow$\;
         \pushline global\_submap\_bounding\_boxes.Intersect(\;
        \pushline\dosemic current\_bounding\_boxes[$si$])\;\popline\popline
  \lnl{algline:local_frontier_detection_begin} Threshold and classify cells in $si$ according to \eqref{eq:thresholding}\;
  \lnl{algline:edge_detection_for} \ForEach{ cell $(k, l) \in$ submap $si$}{
    \lnl{algline:edge_detection} \If{ \nosemic cell $(k, l)$ is \textbf{unobserved} $\wedge$ \;
          cell $(k, l)$ is adjacent to a \textbf{unoccupied} cell in $si$}{
         \lnl{algline:edge_detection_end} local\_frontiers[$si$].Add($(k, l)$)\;
         \lnl{algline:local_frontier_detection_end} TestAndAddToGlobalFrontier()\;
        }
  }
   \lnl{algline:test_adjacent_global_frontiers_against_active_submaps_begin} \ForEach{ submap $sj \in$ intersecting\_submaps}{
     \nl \ForEach{ global\_frontier\_point $\in$ global\_frontiers[$sj$] } {
          \lnl{algline:test_adjacent_global_frontiers_against_active_submaps_end} \If{ \nosemic \textbf{not} StabbingQueryTest(\;
          \pushline\pushline global\_frontier\_point, $si$)\;
          \popline\popline}{
          \lnl{algline:remove_adjacent_global_frontiers} \nosemic global\_frontiers[$sj$].Remove(\;
          \dosemic\pushline global\_frontier\_point)\;\popline
          \lnl{algline:mark_adjacent_global_frontiers_changed} submaps\_with\_frontier\_updates.Add($sj$)\;
      }
     }
   }
 }
 \lnl{algline:begin_insert_finished_submap_bounding_boxes} \ForEach{submap $si \in$ active\_submaps}{
   \nl \If{submap $si$ has just been finished}{
      \lnl{algline:end_insert_finished_submap_bounding_boxes} \nosemic global\_submap\_bounding\_boxes.Insert(\;
      \pushline\dosemic current\_bounding\_boxes[$si$])\;\popline
    }
  }
  \nl \ForEach{ \nosemic submap $si \in$ submaps\_with\_frontier\_updates}{
    \lnl{algline:publishing_of_incremental_frontier_updates} PublishFrontierUpdates($si$, global\_frontiers[$si$])\;
  }
\caption{Handling of updates to active submaps}
\label{alg:handle_submap_update}
\end{algorithm}

\removelatexerror
\begin{algorithm}[H]
    \small
    \capstart
    \KwIn{all variables from call site, by reference}
    \lnl{algline:proc_transforming_points} \nosemic point\_transformed\_to\_global $\Leftarrow$\;
    \pushline \dosemic TransformToGlobal(global\_submap\_poses[$si$],
    cell $(k, l)$)\;\popline
    \lnl{algline:hint_comment} \tcp{Out of all submaps from the set to test} \tcp{against, first try testing against the} \tcp{failing submap hint, if present.}
    \lnl{algline:proc_testing} \eIf{
    \nosemic StabbingQueryTest(\; 
    \pushline \quad point\_transformed\_to\_global,\;
    \quad\quad\,\enspace intersecting\_submaps $\cup$ active\_submaps $\setminus$ $si$)
    \popline
    }{
      \lnl{algline:proc_insert_into_global_frontier} global\_frontiers[$si$].Add(point\_transformed\_to\_global)\;
    }{
      \lnl{algline:proc_add_failing_submap_hint} local\_frontiers[$si$].AddFailingSubmapHintForCell($(k, l)$)\;
    }
\caption{TestAndAddToGlobalFrontier procedure}
\label{alg:testandaddtoglobalfrontier_proc}
\end{algorithm}
\end{figure}
\capstarttrue

\subsection{The frontier detection algorithm}

There are two kinds of events which occur during SLAM execution which are of interest for frontier detection: a submap update event, where a scan is inserted into the active submaps; and a pose graph optimization event which also occurs periodically, but less often.

\autoref{alg:handle_submap_update} describes handling of submap update events, while \autoref{alg:handle_optimization_update} describes handling of pose graph optimization events.

\subsubsection{Handling of submap updates}

For each laser scan processed in SLAM, both submaps in the active pair are updated, making submap updates the most frequently occurring type of event. Because a submap update can only affect the frontier in the area covered by the active submaps, the frontier detection algorithm can be constrained to this area in order to maximize efficiency.

The first step in handling a submap update of an active submap is performing occupancy grid classification and dense local frontier detection on the new version of the submap occupancy grid (\autoref{alg:handle_submap_update}, \crefrange{algline:local_frontier_detection_begin}{algline:local_frontier_detection_end}). In other words, on the local submap level, we have opted to perform a naive edge detection approach instead of e.g. one of the more elaborate approaches described in \autoref{sec:related_work}.

The reason for using a naive approach for local frontier detection is twofold. First, because the submaps are bounded in size (controlled by the fixed parameter $n_{scans}$), and the number of active submaps is constant (2 active submaps per active robot in a multi-robot use case), the time complexity of local frontier detection is not affected by the size of the global map or the size of the dataset. Second, probability thresholding, classification (\autoref{algline:local_frontier_detection_begin}) and edge detection (\crefrange{algline:edge_detection_for}{algline:edge_detection_end}) can be vectorized, allowing for high performance on modern CPUs. Our implementation relies on Eigen \cite{eigenweb}, where we use Eigen's matrix block algebra and Hadamard products to vectorize thresholding, classification and Boolean logic for edge detection.

Computing and storing the set of local frontier points for an active submap produces a set of \emph{candidates} for the global frontier. Every local frontier point is transformed into the global map coordinate system $g$ according to the current global pose of the corresponding submap (\autoref{alg:testandaddtoglobalfrontier_proc}, \autoref{algline:proc_transforming_points}) and the stabbing query test is performed against the intersecting submaps (\autoref{alg:testandaddtoglobalfrontier_proc}, \autoref{algline:proc_testing}). In case the test passes against all submaps, the transformed frontier point is indeed a global frontier point and is added to the global frontier set (\autoref{alg:testandaddtoglobalfrontier_proc}, \autoref{algline:proc_insert_into_global_frontier}). If the test fails, the submap against which the test failed is recorded as a hint (\autoref{alg:testandaddtoglobalfrontier_proc}, \autoref{algline:proc_add_failing_submap_hint}) so that future re-testing (in \autoref{alg:handle_optimization_update}) can be performed faster.

In the next step of handling submap updates, we exploit several properties of graph SLAM based on occupancy grid submaps: 
\begin{enumerate}
  \item The occupancy grids of finished submaps are immutable. Therefore, it is not necessary to re-detect \emph{local} frontiers for already finished submaps.
  \item The algorithm for handling submap updates may assume that no graph optimization has occurred since the last submap update event, so all existing \emph{global} frontiers of finished submaps are valid (except for the situation described below).
  \item The cell occupancy probabilities of active submaps have decreasing entropy, which means that only previously \textbf{unobserved} cells can become \textbf{observed}, and not vice-versa. In other words, updates to active submaps can invalidate (cover up) the previously valid global frontiers of intersecting submaps.
\end{enumerate}

In \autoref{alg:handle_submap_update}, \crefrange{algline:test_adjacent_global_frontiers_against_active_submaps_begin}{algline:test_adjacent_global_frontiers_against_active_submaps_end}, it is tested if the new versions of active submaps cover up existing valid global frontiers of the intersecting finished submaps. This is done by performing stabbing query tests of the global frontier points against the active submaps.
If the stabbing query test fails, the newly covered up global frontier points are removed from the set of global frontier points (\autoref{algline:remove_adjacent_global_frontiers}), thus preserving the invariant that all global frontier points at a time are valid. The global frontiers of finished submaps whose global frontier points got removed are also marked as updated (\autoref{algline:mark_adjacent_global_frontiers_changed}) for incremental publishing of frontier updates (\autoref{algline:publishing_of_incremental_frontier_updates}).

Bounding boxes of finished submaps are stored inside a spatial index tree data structure which enables fast queries of submaps which intersect with a given bounding box (used for looking up finished submaps intersecting with an active submap in \autoref{alg:handle_submap_update}, \autoref{algline:intersecting_submaps_query}). When a submap is marked as finished, its bounding box is inserted into the tree structure (\crefrange{algline:begin_insert_finished_submap_bounding_boxes}{algline:end_insert_finished_submap_bounding_boxes}). Our implementation uses the Boost implementation of R-trees \cite{boost_rtree} for storing global axis-aligned bounding boxes of finished submaps.

It can also be noted that not all submap update events have to be handled in order to guarantee a correct result -- any non-final submap update can be skipped. A robotic exploration system which does not require real-time frontier updates after every scan, but rather occasionally, can invoke the algorithm for handling submap updates only when a submap is finished. This can significantly reduce the computational effort of keeping the frontier up to date, up to a factor of $n_{scans}$.

\capstartfalse
\begin{figure}
\removelatexerror
\begin{algorithm}[H]
 \small
 \capstart
 \KwIn{\nosemic updated global\_submap\_poses, \; \quad finished\_submaps, active\_submaps}
 \KwPersistent{\\ \quad local\_frontiers, global\_frontiers, \\ \quad global\_submap\_bounding\_boxes}
\lnl{algline:bounding_boxes_recomputation_begin} global\_submap\_bounding\_boxes.Clear()\;
\nl \ForEach{submap $si \in$ finished\_submaps}{
\lnl{algline:bounding_boxes_recomputation_end} \nosemic global\_submap\_bounding\_boxes.Insert(\;
   \pushline CalculateGlobalBoundingBox($si$,\;
        \pushline\dosemic global\_submap\_poses[$si$]))\;\popline
}
\nl global\_frontiers.Clear()\;
\nl \ForEach{submap $si \in$ finished\_submaps $\cup$ active\_submaps}{
   \lnl{algline:alg2_intersecting_submaps} \nosemic intersecting\_submaps $\Leftarrow$\;
      \Indp global\_submap\_bounding\_boxes.Intersect(\;
      \Indp global\_submap\_bounding\_boxes[$si$])\;
      \dosemic $\cup$ active\_submaps $\setminus$ $si$\Indm\;\Indm
   \nl \ForEach{ cell $(k, l) \in$ local\_frontiers[$si$]}{
      \nl TestAndAddToGlobalFrontier()\;
   }
   \nl PublishFrontierUpdates($si$, global\_frontiers[$si$])\;
}
\caption{Handling of pose graph optimization events}
\label{alg:handle_optimization_update}
\end{algorithm}
\end{figure}
\capstarttrue

\subsubsection{Handling of graph optimization events}

When graph SLAM performs optimization, a new solution is produced for poses of all graph nodes. For frontier detection, this means that the submap poses have changed. This invalidates the global bounding boxes of submaps and the entire global frontier, all of which has to be recomputed in \autoref{alg:handle_optimization_update}. However, advantage is taken of the fact that the local frontiers have already been computed for all submaps, so all that needs to be done is to re-transform the local frontier points to the global coordinate system $g$ and re-test them.

The global bounding boxes of submaps are recomputed in \crefrange{algline:bounding_boxes_recomputation_begin}{algline:bounding_boxes_recomputation_end}, while the rest of \autoref{alg:handle_optimization_update} recomputes the global frontier similarly to \autoref{alg:handle_submap_update} using \autoref{alg:testandaddtoglobalfrontier_proc}: each local frontier point is re-transformed according to the corresponding new global submap pose and re-tested. If the stabbing query test is passed, the frontier point is inserted into the new global frontier set.

During re-testing of a frontier point against all submaps in the set of intersecting submaps, it is advisable to first try testing against the submap stored in the failing submap hint, if it exists (comment in \autoref{alg:testandaddtoglobalfrontier_proc}, \autoref{algline:hint_comment}). This speeds up rejection of points which fail the test by first testing against the same submap which caused the local frontier point to fail the test earlier. 

Recomputing the entire global frontier in \autoref{alg:handle_optimization_update} is  not a computationally lightweight operation. However, it is more efficient than a non-submap aware approach that would require iterating through the reassembled global map with time complexity proportional to the \emph{area} of the global map. Recomputing the global frontier in \autoref{alg:handle_optimization_update} has the time complexity proportional to the \emph{perimeter} of local frontiers.

To summarize, \autoref{alg:handle_submap_update} assumed that all global frontier points before a submap update event were valid and up-to-date. Graph optimization violates this assumption by displacing the submaps according to submap poses of the new pose graph solution. \autoref{alg:handle_optimization_update} restores this invariant by efficiently recomputing the global frontier.

\section{Algorithm analysis}

\subsection{Soundness and completeness}

Let us first note that the process of merging submaps into a global map according to a pose graph solution cannot result in an unobserved global map cell if any of the corresponding submap cells are observed. In other words, if a global map cell is unobserved, the corresponding cells in all submaps are also unobserved. Also, let $LF$ and $GF$ denote the sets of local frontier points of all submaps (transformed to $g$, but untested) and all valid global frontier points, respectively.

For discussing completeness, we will consider a valid global frontier point in $GF$, i.e. the center of an \emph{unobserved} global map cell adjacent to a \emph{unoccupied} global map cell. Due to the map merging process, all submap cells corresponding to the global frontier cell have to be unobserved as well. Next, the adjacent unoccupied global map cell is also unoccupied in at least one submap, whose merging caused the unoccupied global map cell to be marked as such. In that submap, the unobserved cell next to the unoccupied cell is a local frontier and is thus in $LF$. Therefore, each valid global frontier point corresponds to a local frontier point in at least one submap (i.e. $GF \subseteq LF$), and an algorithm which computes $GF$ by taking $LF$ as input is thus complete, providing it does not incorrectly discard any valid global frontier points in the process. 

The rule for discarding local frontier points when they fail the stabbing query test can only result in \emph{observed} cells being correctly discarded from the global frontier, and therefore the property of completeness is preserved. Also, the naive edge detection algorithm used for computing the local frontiers is trivially valid.

For soundness, we will consider a global frontier point returned by the proposed algorithm, and suppose that it is not valid. This could be either because the returned global frontier point is an observed global map cell, or because the returned global frontier point is not adjacent to a unoccupied global map cell. 
The first case is not possible because the frontier point would have failed the stabbing query test against the submap that contains an observed cell which resulted in marking the corresponding global map cell as observed. 

The second case is possible but unlikely -- for example, if there were a plurality of submaps with corresponding adjacent cells being occupied rather than unoccupied, so the merging process results in adjacent global map cells being marked as occupied instead of unoccupied, and the global frontier point is adjacent only to occupied global map cells. 
This is unlikely because the chance of the unobserved cell not getting covered up by observed cells (which in turn will make it fail the stabbing query test) is negligible. To exclude this case and to make the algorithm completely sound, the stabbing query test could be modified to also look at the adjacent cells in submaps as well, instead of just checking single cells. 
This would not increase the theoretical time complexity, but would make the implementation unnecessarily more elaborate and slow.

\subsection{Computational complexity}

\subsubsection{Handling of submap updates}

In \autoref{alg:handle_submap_update}, \autoref{algline:intersecting_submaps_query}, the complexity of querying the R-tree for submaps intersecting with the updated submap $si$ is $O(\log|S| + |S_{\cap si}|)$, where $S$ is the set of all submaps, and $S_{\cap si}$ is the set of submaps intersecting with the submap $si$. This also includes the complexity of inserting a finished submap bounding box into the R-tree.

In \crefrange{algline:local_frontier_detection_begin}{algline:local_frontier_detection_end}, occupancy grid classification and naive local frontier detection run in $O(A(si))$, where $A(si)$ is the area of submap $si$, i.e. the number of cells in the submap.

The number of local frontier points in $si$, i.e. the perimeter of its local frontier $LF_{si}$ will be denoted as $P(LF_{si})$, while the global frontier of the submap $si$ and its perimeter will be denoted as $GF_{si}$ and $P(GF_{si})$, respectively.

For each detected local frontier point in the updated submap, the complexity of performing the stabbing query test against the intersecting submaps (\autoref{alg:testandaddtoglobalfrontier_proc}, \autoref{algline:proc_testing}) is $O(|S_{\cap si}|)$, yielding the total complexity of this step $O(P(LF_{si}) \cdot |S_{\cap si}|)$.

Invalidating global frontiers of the intersecting submaps (\crefrange{algline:test_adjacent_global_frontiers_against_active_submaps_begin}{algline:mark_adjacent_global_frontiers_changed}) runs linearly with respect to their perimeter, yielding complexity $O(P(\bigcup_{sj \in S_{\cap si}} GF_{sj}))$.

The total time complexity of handling an update of submap $si$ thus equals 
(simplified assuming $P(LF_{si}) \geq 1$):
\begin{equation}
O(\log|S| + A(si) + P(LF_{si}) \cdot |S_{\cap si}| + P(\bigcup_{sj \in S_{\cap si}} GF_{sj}))
\end{equation}

\begin{table}[t]
\centering
\caption{Experimental setup and results}
\label{table:results}
\begin{tabular}{l}
\toprule
\textbf{Testing setup details} \\
   \quad Intel i7 6800K @ 3600 MHz, Ubuntu 18.04, ROS Melodic\\
   \quad Cartographer offline node, exit before final optimization \\
   \quad No RViz visualization \\ \midrule
\textbf{Dataset}: \href{https://storage.googleapis.com/cartographer-public-data/bags/backpack_2d/cartographer_paper_deutsches_museum.bag}{cartographer\_paper\_deutsches\_museum.bag}\\
   \quad Undisclosed IMU and multi-echo 2D lidar (scan rate: $\sim$36 Hz) \\
   \quad Duration: 1912 s \quad Covered area: $\sim$12 000 m$^2$ \quad Path length: 2272 m\\
   \quad Differences from default SLAM parameters provided by Google: \\
   \quad \quad POSE\_GRAPH.constraint\_builder.sampling\_ratio = 0.1\\ 
   \quad \quad MAP\_BUILDER.num\_background\_threads = 10 \\
\textbf{Wall clock SLAM processing time} (not including final optimization) \\
        \quad without frontier detection: 360 s (5.3x realtime on average) \\
        \quad with asynchronous frontier detection: 434 s (4.4x realtime on average) \\
        \quad with synchronous frontier detection: 605 s (3.2x realtime on average) \\
\textbf{Wall frequency of asynchronous frontier updates} \\ 
  \quad Average: \textbf{78 Hz} (13 ms)
  \quad Std. deviation: 22 ms\\
\textbf{Average synchronous frontier update processing wall time} \\ 
  \quad (605 s $\pm$ 4.5 s - 360 s) / 37816 scans = \textbf{6.5 ms per scan} $\pm$ 0.12 ms  \\
  \textbf{SLAM events} \\
   \quad Total submap update events (inserted laser scans): 37816 \\
   \quad Skipped submap update events (asynchronous only): 3900/37816\\
   \quad Optimization events handled during bag processing: \\
   \quad \quad Asynchronous: 355 out of 420 \quad Synchronous: 412 out of 420\\
   \quad Average wall time between optimization events: \\
   \quad \quad Asynchronous: 1.22 s \quad\quad\quad\quad\quad Synchronous: 0.68 s \\ \midrule
\textbf{Dataset}: \href{https://github.com/larics/cartographer_frontier_detection/tree/master/fr079-dataset}{Freiburg FR-079 uncorrected} \\
  \quad SICK LMS 2D lidar (scan rate: 4.7 Hz) \\
  \quad Duration: 1061 s \quad Covered area: $\sim$90 m$^2$ \quad \quad Path length: 396 m \\
\textbf{Wall clock SLAM processing time} (not including final optimization) \\ 
        \quad without frontier detection: 16.5 s (64x realtime on average) \\
        \quad with synchronous frontier detection: 20.3 s (52x realtime on average) \\
\textbf{Average synchronous frontier update processing wall time} \\ 
  \quad (20.3 s $\pm$ 0.3 s - 16.5 s) / 4198 scans = \textbf{0.92 ms per scan} $\pm$ 0.08 ms \\
  \textbf{SLAM events} \\
   \quad Total submap update events (inserted laser scans): 4198 \\
   \quad Skipped submap update events: none (synchronous) \\
   \quad Optimization events handled during bag processing: 25 out of 27\\
   \quad Average wall time between optimization events: 0.82 s\\ \bottomrule
\end{tabular}
\end{table}

\subsubsection{Handling of pose graph optimization events}

The first step in \autoref{alg:handle_optimization_update} is to rebuilt the R-tree (\crefrange{algline:bounding_boxes_recomputation_begin}{algline:bounding_boxes_recomputation_end}), which runs in $O(|S|\log|S|)$. This complexity also includes looking up the intersecting submaps for each submap (\autoref{algline:alg2_intersecting_submaps}).

Next, every local frontier point is transformed to the global coordinate system $g$ and the stabbing query test is performed against the intersecting submaps (\autoref{alg:testandaddtoglobalfrontier_proc}, \crefrange{algline:proc_transforming_points}{algline:proc_testing}). A pessimistic bound would entail testing every local frontier point against all other submaps, i.e. a time complexity of $O(|S| \cdot P(LF))$. The pessimism of this bound can be reduced by assuming that the points which fail the stabbing query test will do so on the first test, performed against the submap stored in the failing submap hint. If the perimeter of points which fail the test, equal to $P(LF) - P(GF)$, is denoted as $P(FF)$, this assumption yields the time complexity of $O(|S|\cdot P(GF) + P(FF))$.

\setlength\abovedisplayskip{3pt}
The total time complexity of handling a pose graph optimization event thus depends on the number of submaps and the local and global frontier perimeters:
\begin{equation}
O(|S| \cdot (\log|S| + P(GF)) + P(FF))
\end{equation}

We have managed to avoid having the two-dimensional map area in the time complexity of the global operation of handling pose graph optimization by taking advantage of submaps and their immutability.

\section{Experimental results} \label{sec:experimental_results}

Our implementation of the proposed frontier detection algorithm (available on Github\footnote{\href{https://github.com/larics/cartographer_frontier_detection}{https://github.com/larics/cartographer\_frontier\_detection}}) has been developed as an extension of Cartographer. We have chosen to focus our experimental validation on running offline SLAM processing with frontier detection on a single-robot, but otherwise quite demanding publicly available dataset: the introductory example for Cartographer, the Google Deutsches Museum bag \cite{hess2016}, captured with a human-carried sensor backpack. 

We have also performed an evaluation with the Freiburg FR-079 dataset\footnote{ROS bag and SLAM parameters are available in the Github repository} in order to roughly compare with \cite{quin2014}, who tested most of the algorithms described in \autoref{sec:related_work} in a simulated FR-079 environment.

Demo videos of frontier detection for both evaluated datasets are available\footnote{\href{https://goo.gl/62zEUy}{https://goo.gl/62zEUy}}, along with an additional demo of a multi-robot scenario where the dataset was recorded with two teleoperated Pioneer mobile robots driven side-by-side.\footnote{Dataset and SLAM parameters also available in the Github repository}

We have opted to use the Google Cartographer \emph{offline node}, which processes a ROS bag dataset as fast as the CPU can handle (around 4-5x realtime on Deutsches Museum), enabling us to demonstrate frontier update frequencies far greater than the scan frequency of a single robot's LIDAR. 

In order to minimize the impact on SLAM performance, the frontier detection algorithm may execute asynchronously, in a separate thread. The frontier detection algorithm tries to process all submap updates, while it can adaptively skip non-final submap updates in case the processing speed falls behind SLAM. In the asynchronous case on Deutsches Museum, the wall clock frequency of incremental frontier updates has been measured as a benchmark of frontier detection performance.

Running frontier detection synchronously (in the main SLAM thread) makes it easier to measure the exact time spent performing frontier detection, which can be measured as the difference of wall clock durations of processing a dataset with synchronous frontier detection and with no frontier detection at all.

The results are given in \autoref{table:results}. We believe that the Deutsches Museum result is an achievement with respect to the state of the art. To compare, the best performing-algorithm in \cite{quin2014}, EWFD, achieved 290 ms and 190 ms per frontier update iteration in two simulated FR-097 environment trajectories, with significantly lower numbers of processed scans (252 scans for the 290 ms \emph{Freiburg 1}). We have achieved sub-millisecond (0.92 ms) wall-time per frontier update, while handling a fairly larger number of scans (4198) and 25 pose graph optimizations on average. We do have to point out that our experiments have been performed on newer hardware, and that we have not performed frontier grouping, but rather return the detected frontier as an unstructured set of points.


\subsection{Additional notes on implementation of the frontier detection algorithm}

An optimization that we have implemented is performing additional stabbing query tests against a small number of \emph{temporally close} submaps 
in the local frontier edge detection condition in \autoref{alg:handle_submap_update}, \autoref{algline:edge_detection} (for example, against 4 previous submaps). This has the result of permanently ``baking in'' a negative test result against these submaps, since these points are permanently erased from the local frontiers. 

A benefit of discarding points early from local frontiers is speeding up recomputation of the global frontier in \autoref{alg:handle_optimization_update}, because there will be fewer candidates in the local frontier which have to be re-transformed and re-tested. 
Also, it is expected that the localization drift between sequential submaps is small, and that pose graph optimization will not produce a significant relative displacement between sequential submaps. Using the unoptimized poses may actually be preferential for processing sequential submaps, because it could prevent detecting a false frontier resulting from slight misalignment of sequential submaps introduced by pose graph optimization when closing loops.

We have employed a few cosmetic improvements which result in (subjectively) aesthetically better frontiers. The first is a change to \eqref{eq:thresholding} where cells with a very uncertain ``unoccupied'' probability (i.e. $\mat{S}^{si}_{k, l} \in [0.5 - \varepsilon, 0.5]$, where $\varepsilon \colonequals 0.04$ is an arbitrarily chosen small value) are treated as ``unobserved'' cells. This prevents detecting a false frontier around single false long-distance laser readings, which cause insertion of a false ray of ``unoccupied'' cells into the submap. This will also have the beneficial effect of driving the robot exploration system to get a ``better look'' at areas with uncertain unoccupied cells, since they are considered unexplored.

The second change we have made is simple smoothing of local frontiers by adjusting the definition of a local frontier to be \emph{the center of an unobserved cell with $\geq$ 2 unoccupied adjacent cells and $\geq$ 2 unobserved adjacent cells}, where \emph{adjacent} cells are cells in the Moore 8-neighbourhood. With no other changes to the rest of the algorithm, these changes result in a smoother global frontier, since the smoothed local frontier points are candidates for the global frontier. 

\section{Conclusion and future work}

We have described, implemented and tested an efficient frontier detection algorithm that is specialized for 2D multi-robot graph SLAM based on occupancy grid submaps. Our algorithm is efficiently constrained to the area of active submaps, yet robust to loop closure by efficiently recomputing the global frontier after pose graph optimization is performed without the time complexity being a function of map area.

\emph{Future work:}
Using the proposed frontier detection algorithm in an actual closed exploration-system loop (\autoref{fig:high_level_block_diagram}) is a work in progress.
Also, some cited state of the art (e.g. \cite{keidar2014}) performs grouping of continuous frontier points into segments, which is useful for selecting navigation objectives of exploration tasks. Another kind of post-processing of the detected frontier which might be of interest is reachability analysis, ie. detecting frontier points such as points behind glass, closed doors or behind walls (e.g. from false laser readings). Nonetheless, a performant algorithm for frontier detection is the basis for any such further improvements.

\bibliographystyle{IEEEtran}
\bibliography{frontier}
%

\end{document}